\documentclass[10pt,twocolumn,letterpaper]{article}

\usepackage{icb}
\usepackage{times}
\usepackage{epsfig}
\usepackage{graphicx}
\usepackage{amsmath}
\usepackage{amssymb}
\usepackage{subfigure}

\usepackage{color}

\icbfinalcopy % *** Uncomment this line for the final submission

\ificbfinal\fi
\begin{document}

%%%%%%%%% TITLE
\title{Kinship Verification from Videos using Spatio-Temporal Texture Features and Deep Learning}

\author{Elhocine Boutellaa$^{1,2,3}$, Miguel Bordallo L\'opez$^{2}$, Samy Ait-Aoudia$^{3}$, Xiaoyi Feng$^{4}$, and Abdenour Hadid$^{2,4}$\\
$^1$Centre de D\'eveloppement des Technologies Avanc\'ees, Algeria \\
$^2$University of Oulu, Finland\\
$^3$Ecole Nationale Sup\`erieure d'Informatique, Algeria\\
$^4$Northwestern Polytechnical University, China
%\and
}

\maketitle
\thispagestyle{empty}

%---------------------------------------------------------------------------------------------------------------------------
\begin{abstract}
Automatic kinship verification using facial images is a relatively new and challenging research problem in computer vision. It consists in automatically predicting whether two persons have a biological kin relation by examining their facial attributes. While most of the existing works extract shallow handcrafted features from still face images, we approach this problem from spatio-temporal point of view and explore the use of both shallow texture features and deep features for characterizing faces. Promising results, especially those of deep features, are obtained on the benchmark UvA-NEMO Smile database. Our extensive experiments also show the superiority of using videos over still images, hence pointing out the important role of facial dynamics in kinship verification. Furthermore, the fusion of the two types of features (i.e. shallow spatio-temporal texture features and deep features) shows significant performance improvements compared to state-of-the-art methods.
\end{abstract}
%---------------------------------------------------------------------------------------------------------------------------
\vspace{-1cm}
\section{Introduction}
It is a common and an easy practice for us, humans, to identify our relatives from faces. Relatives usually wonder which facial attributes does a new born baby inherit from which family member. The human ability of kinship recognition has been the object of many psychological studies \cite{dal2006kin,DeBruine_VR09}. Inspired by these studies,  automatic kinship (or family) verification  \cite{Fang_icip10:1stKin,Zhou_mm11:SPLE} has been recently considered as an interesting and open research problem in computer vision which is receiving an increasing attention from the research community. %\looseness=-1

Automatic kinship verification from faces aims at determining whether two persons have a biological kin relation or not by comparing their facial attributes. Kinship verification is important for automatically analyzing the huge amount of photos daily shared on social media. It helps understanding the family relationships in these photos. Kinship verification is also useful in case of missing children, elderly people with Alzheimer or possible kidnapping cases. For instance, a suspicious behavior between two persons (e.g. an adult and a child) captured by a surveillance camera can be subject to further analysis to determine whether they are from the same family or not to prevent crimes and kidnapping. Kinship verification can also be used for automatically organizing family albums and generating family trees.

Kinship verification using only facial images is a very challenging task. It inherits the research problems of face verification from images captured in the wild under adverse pose, expression, illumination and occlusion conditions. In addition, kinship verification should deal with wider intra-class and inter-class variations, as persons from the same family may look very different while faces of persons with no kin relation may look similar. Moreover, automatic kinship verification poses new challenges, since a pair of input images may be from persons of different sex (e.g. brother-sister kin) and/or with a large age difference (e.g. father-daughter kin).

The published papers and organized competitions (e.g.~\cite{lu_2014:1Comp,Lu_FG15:comp}) dealing with automatic kinship verification over the past few years have shown some promising results. Typical current best-performing methods combine several face descriptors, apply metric learning approaches and compute Euclidean distances between pairs of features for kinship verification. It appears that most of these works are mainly based on shallow handcrafted features. Hence, they are not associated with the recent significant progress in machine learning, that suggests the use of deep features. Moreover, the role of facial dynamics in kinship verification is mostly unexplored as allmost all the existing works focus on analyzing still facial images instead of video sequences. Based on these observations, we propose to approach the problem of kinship verification from a spatio-temporal point of view and to exploit the recent progress in deep learning for facial analysis.

Given two face video sequences, to verify their kin relationship, our proposed approach starts with detecting, segmenting and aligning the face images based on eye coordinates. Then, two types of descriptors are extracted: shallow spatio-temporal texture features and deep features. As spatio-temporal features, we extract local binary patterns (LBP)~\cite{Ahonen:lbp_face}, local phase quantization (LPQ)~\cite{lpq} and binarized statistical image features (BSIF)~\cite{bsif_icpr12}. These features are all extracted from Three Orthogonal Planes (TOP) of the videos. Deep features are extracted by convolutional neural networks (CNNs)~\cite{Parkhi15}. The feature vectors of face pairs to compare are then combined to be used as inputs to Support Vector Machines (SVM) for classification.
We conduct extensive experiments on the benchmark UvA-NEMO Smile database~\cite{Dibeklioglu_eccv12} obtaining very promising results, especially with the deep features. The results also clearly demonstrate the superiority of using videos over still images, hence pointing out the important role of facial dynamics in kinship verification. Furthermore, the fusion of the two types of features (i.e. shallow spatio-temporal texture features and deep features) results in significant performance improvements compared to state-of-the-art methods.

The rest of the paper is organized as follows. Section~\ref{relWork} discusses the related work. Our proposed approach for kinship verification from videos is described in Section~\ref{proposed}. Experiments and analysis are shown in Section~\ref{exp}. Section~\ref{conc} draws conclusions and points out future research directions.
%--------------------------------------------------------------------------------------------------------------------------------
\section{Related work}\label{relWork}
Kinship verification using facial images is receiving increasing interest from the research community. This is mainly motivated by its potential applications, especially in analyzing daily shared data in social web. The first approaches to tackle kinship verification were based on low-level handcrafted feature extraction and SVM or \textit{K}-NN classifiers. For instance, Gabor gradient orientation pyramids have been used by~ Zhou~\textit{et~al.} \cite{Zhou_MM12:Gabor_pyramid}; Yan~\textit{et~al.}~\cite{Zhou_mm11:SPLE} used a spatial pyramid learning descriptor; and Self-similarity of Weber faces is used by Kohli~\textit{et~al.}~\cite{Kohli_BTAS12:weber}. However, the best performance is usually obtained by combining several types of features. For example, in the last kinship competition~\cite{Lu_FG15:comp}, all the proposed methods used three or more descriptors. The best performing method in this competition employed four different local features (LBP, HOG, OCLBP and Fisher vectors). 

On the other hand, different metric learning approaches have been investigated to tackle the kinship verification problem. For example, Lu~\textit{et~al.}~\cite{Lu_PAMI14:NRML} learned a distance metric where the face pairs with a kin relation are pulled close to each other and those without a kin relation are pushed away. Recently, Zhou~\textit{et~al.}~\cite{Zhou_IF15} applied ensemble similarity learning for solving the kinship verification problem. They learned an ensemble of sparse bi-linear similarity bases from kinship data by minimizing the violation of the kinship constraints between pairs of images and maximizing the diversity of the similarity bases. Yan~\textit{et~al.}~\cite{yan_IFS14} and Hu~\textit{et~al.}~\cite{hu_ACCV14:LMML} learned multiple distance metrics based on various features, by simultaneously maximizing the kinship constraint (pairs with a kinship relation must have a smaller distance than pairs without a kinship relation) and the correlation of different features.

The most recent trends are motivated by the impressive success of deep learning approaches in various image representation and classification~\cite{Simonyan_arxiv14} in general and face recognition in particular~\cite{Sun_cvpr2014}. Zhang~\textit{et~al.}~\cite{Zhang_BMVC15:deep} recently proposed a convolution neural network architecture for face-based kinship verification. The proposed architecture is composed by two \textit{convolution max pooling} layers followed by a \textit{convolution} layer then a \textit{fully connected} layer. A two-way \textit{softmax} classifier is used as the final layer to train the network. The network takes a pair of RGB face images of different persons as an input, checking the possible kin relations. However, their reported results do not outperform the shallow methods presented in the FG15 kinship competition on the same datasets~\cite{Lu_FG15:comp}. The reason behind this may be the scarcity of training data, since deep learning approaches require the availability of enough training samples, a case that is not fulfilled by the currently available face kinship databases. 

While most of the published works cope with kinship problem from images, to our knowledge the only work that performed kinship from videos was conducted by Dibeklioglu~\textit{et~al.}~\cite{Dibeklioglu_ICCV13}. The authors combined facial expression dynamics with temporal facial appearance as features and used SVM for classification. In the present work, we aim to exploit the temporal information present in face videos, investigating the use of both spatio-temporal shallow features and deep features describing faces.
%-------------------------------------------------------------------------------------------------------------------------
%----------------------VGG-Face Table-------------------------------------
\begin{table*}[htb!]
\resizebox{\textwidth}{!}{
\begin{tabular}{|l|ccccccccccccccccccc|}
\hline
layer&0&1&2&3&4&5&6&7&8&9&10&11&12&13&14&15&16&17&18 \\ 
type&input&conv&relu&conv&relu&mpool&conv&relu&conv&relu&mpool&conv&relu&conv&relu&conv&relu&mpool&conv \\ 
name&–&conv1\_1&relu1\_1&conv1\_2&relu1\_2&pool1&conv2\_1&relu2\_1&conv2\_2&relu2\_2&pool2&conv3\_1&relu3\_1&conv3\_2&relu3\_2&conv3\_3&relu3\_3&pool3&conv4\_1 \\ \hline %\hline
support&–&3&1&3&1&2&3&1&3&1&2&3&1&3&1&3&1&2&3 \\ 
filt dim&–&3&–&64&–&–&64&–&128&–&–&128&–&256&–&256&–&–&256 \\ 
num filts&–&64&–&64&–&–&128&–&128&–&–&256&–&256&–&256&–&–&512 \\ 
stride&–&1&1&1&1&2&1&1&1&1&2&1&1&1&1&1&1&2&1 \\ 
pad&–&1&0&1&0&0&1&0&1&0&0&1&0&1&0&1&0&0&1 \\ \hline \hline
%---------------------------------------------------------------
layer&19&20&21&22&23&24&25&26&27&28&29&30&31&32&33&34&35&36&37 \\
type&relu&conv&relu&conv&relu&mpool&conv&relu&conv&relu&conv&relu&mpool&conv&relu&conv&relu&conv&softmx \\
name& relu4\_1&conv4\_2&relu4\_2&conv4\_3&relu4\_3&pool4&conv5\_1&relu5\_1&conv5\_2&relu5\_2&conv5\_3&relu5\_3&pool5&fc6&relu6&fc7&relu7&fc8&prob \\
support&1&3&1&3&1&2&3&1&3&1&3&1&2&7&1&1&1&1&1 \\
filt dim&–&512&–&512&–&–&512&–&512&–&512&–&–&512&–&4096&–&4096&– \\
num filts&–&512&–&512&–&–&512&–&512&–&512&–&–&4096&–&4096&–&2622&– \\
stride&1&1&1&1&1&2&1&1&1&1&1&1&2&1&1&1&1&1&1 \\
pad&0&1&0&1&0&0&1&0&1&0&1&0&0&0&0&0&0&0&0 \\ \hline
\end{tabular}
}
%----------------------------------------------
\vspace{.05cm}
\caption{VGG-face CNN architecture.}
\label{tab:vgg-face}
\end{table*}
\vspace{-.2cm}
%-----------------------------------------------------------
\section{Video-based kinship verification}\label{proposed}
This section describes our approach for kinship verification using facial videos. In the following, the steps of the proposed approach are detailed.

\subsection{Face detection and cropping} 
In our approach, the first step consists in segmenting the face region from each video sequence. For that purpose, we have employed an active shape model (ASM) based approach that detects 68 facial landmarks. The regions containing faces are then cropped from every frame in the video using the detected landmarks. Finally, The face-regions are aligned using key landmark points and registered to a predefined template.
%%%%%%%%%%%%%%%%%%%%%%%%%%%%%%%%%%%%%%%%%----------------------------------
\subsection{Face description} 
For describing faces from videos, we use two types of features: texture spatio-temporal features and deep learning features. These features are introduced in this subsection.
%\vspace{-.3cm}
\subsubsection{Spatio-temporal features}
Spatio-temporal texture features have been shown to be efficient for describing faces in various face analysis tasks, such as face recognition and facial expression classification. In this work, we extract three local texture descriptors: LBP~\cite{Ahonen:lbp_face}, LPQ~\cite{lpq} and BSIF~\cite{bsif_icpr12}. These three features are able to describe an image using a histogram of decimal values. The code corresponding to each pixel in the image is computed from a series of binary responses of the pixel neighborhood to a filter bank. In LBP and LPQ the filters are handcrafted while the filters of BSIF are learned from natural images. Specifically, the binary code of a pixel in LBP is computed by thresholding its value with the circularly symmetric $P$ neighboring pixels (on a circle of radius $R$). LPQ encodes the local phase information of four frequencies of the short term Fourier transform (STFT) over a local window of size $W \times W$ surrounding the pixel. BSIF binarizes the responses of $f$ independent filters of size $W \times W$ learnt by independent component analysis (ICA).  

The spatio-temporal textural dynamics of the face in a video are extracted from three orthogonal planes XY, XT, and YT~\cite{lbp-top}, separately. X and Y are the horizontal and vertical spatial axes of the video, and T refers to the time. The texture features of each plane are aggregated into a separate histogram. Then the three histograms are concatenated into a single feature vector. To take benefit of the multi-resolution representation~\cite{mslpq}, the three features are extracted at multiple scales, varying their parameters. For the LBP descriptor, the selected parameters are $P=\{8, 16, 24\}$ and $R=\{1, 2, 3\}$. For LPQ and BSIF descriptors, the filter sizes were selected as $W=\{3,5,7,9,11,13,15,17\}$. 
%%%%%%%%%%%%%%%%%%%%%%%%%%%%%%%%%%%%%%
\subsubsection{Deep learning features}
Deep neural networks have been recently outperforming the state of the art in various classification tasks. Particularly, convolutional neural networks (CNNs) demonstrated impressive performance in object classification in general and face recognition in particular. However, deep neural networks require a huge amount of training data to learn efficient features. Unfortunately, this is not the case for the currently available kinship databases. We conducted preliminary experiments using a Siamese CNN architecture as well as a deep architecture proposed by a previous work \cite{Zhang_BMVC15:deep}. As expected both approaches resulted in lower performance than using shallow features, due to the lack of enough training data.
%%---------------------DB Samples image-------------------------
%\begin{figure*}[htb!]
%\begin{center}
   %\includegraphics[width=.7\linewidth]{dbSamples}
%\end{center}
%%\vspace{-1cm}
   %\caption{Samples of pair images form UvA-NEMO Smile database for different kin relations. Positive pairs are combinations of first row with second row (green rectangles) and negative pairs are combinations of second row with third row (red rectangles). }
%\label{fig:pairsSamples}
%\end{figure*}
%%%----------------------------------------------
An alternative for extracting deep face features is to use a pre-trained network. A number of very deep pre-trained architectures has already been made available to the research community. Motivated by the similarities between face recognition and kinship verification problems, where the goal is to compute the common features in two facial representations, we decided to use the VGG-face~\cite{Parkhi15} network. VGG-face has been initially trained for face recognition on a reasonably large dataset of $2.6$ million images of over $2622$ people. This network has been evaluated for face verification from both pairs of images and videos showing interesting performance compared against state of the art.

The detailed parameters of the VGG-face CNN are provided by Table~\ref{tab:vgg-face}. The input of the network is an RGB face image of size $224\times224$ pixels. The network is composed of $13$ linear convolution layers \textit{(conv)}, each followed by a non-linear rectification layer \textit{(relu)}. Some of these rectification layers are followed by a non-linear max pooling layer \textit{(mpool)}. Following are two fully connected layers \textit{(fc)} both outputting a vector of size $4096$. At the top of the initial network are a \textit{fully connected} layer with the size of classes to predict ($2622$) and a \textit{softmax} layer for computing the class posterior probabilities.

In this context, to extract deep face features for kinship verification, we input the video frames one by one to the CNN and collect the feature vector issued by the fully connected layer fc7 (all the layers of the CNN except the class predictor fc8 layer and the softmax layer are used). Finally, all the frames' features of a given face video are averaged, resulting in a video descriptor that can be used for classification.  
%%%%%%%%%%%%%%%%%%%%%%%%%%%
\subsection{Classification}
To classify a pair of face features as positive (the two persons have a kinship relation) or negative (no kinship relation between the two persons), we use a bi-class linear Support Vector Machine classifier (SVM). Before feeding the features to the SVM, each pair of features has to be transformed into a single feature vector as imposed by the classifier. We have examined various ways for combining a pair of features, such as concatenation and vector distances. We have empirically found that utilizing the normalized absolute difference shows the best performance. Therefore, in our experiments, a pair of feature vectors $X=\{x_1, \dots, x_d\}$ and $Y=\{y_1, \dots, y_d\}$ is represented by the vector $F=\{f_1, \dots, f_d\}$ where :
\begin{equation}
f_i=\sum_j {\frac{\left|x_j-y_j\right|}{\sum_j{ (x_j+y_j)}}}
\end{equation}
%-----------------------------------------------------------------------------------------------------------------------------------
\section{Experiments}\label{exp}
%----------------DB Stat table------------------
\begin{table}
\begin{center}
\begin{tabular}{ | l | c | c | c | c | }
\hline
 \multicolumn{1}{|c|}{}& \multicolumn{2}{|c|}{Spontaneous}   & \multicolumn{2}{|c|}{Posed}  \\ \hline
	Relation & Subj. \# & Vid. \# & Sub. \# & Vid. \# \\ \hline
	S-S & 7 & 22 & 9 & 32 \\ \hline
	B-B & 7 & 15 & 6 & 13 \\ \hline
	S-B & 12 & 32 & 10 & 34 \\ \hline
	M-D & 16 & 57 & 20 & 76 \\ \hline
	M-S & 12 & 36 & 14 & 46 \\ \hline
	F-D & 9 & 28 & 9 & 30 \\ \hline
	F-S & 12 & 38 & 19 & 56 \\ \hline
	All & 75 & 228 & 87 & 287 \\ \hline
\end{tabular}
\end{center}
\vspace{-.3cm}
\caption{Kinship statistics of UvA-NEMO Smile database.}
\label{tab:smiledb}
\end{table} 
%-------------------------------------------
\subsection{Database and test protocol}
%\vspace{-.5cm}
To evaluate the proposed approach, we use UvA-NEMO Smile database~\cite{Dibeklioglu_eccv12}, which is currently the only available video kinship database. The database was initially collected for analyzing posed versus spontaneous smiles of subjects. Videos are recorded with a resolution of $1920\times1080$ pixels at a rate of $50$ frames per second under controlled illumination conditions. A color chart is placed on the background of the videos to allow further illumination and color normalization. The videos are collected in controlled conditions and do not show any kind of bias~\cite{TPAMIours}. The ages of the subjects in the database vary from $8$ to $76$ years. Many families participated in the database collection, allowing its use for evaluation of automatic kinship from videos. A total of $95$ kin relations were identified between $152$ subjects in the database. There are seven different kin relations between pairs of videos: Sister-Sister (S-S), Brother-Brother (B-B), Sister-Brother (S-B), Mother-Daughter (M-D), Mother-Son (M-S), Father-Daughter (F-D), and Father-Son (F-S). The association of the videos of persons having kinship relations gives $228$ pairs of spontaneous and $287$ pairs of posed smile videos. The statistics of the database are summarized in Table~\ref{tab:smiledb}.
%---------------------------------
\begin{table*}[tbh!]
\begin{center}
\begin{tabular}{|l|c|c|c|c|c|c|c||c|c|}
\hline
Method  &    S-S &    B-B & S-B    & M-D    & M-S    & F-D    & F-S    &   Mean & Whole set \\ \hline
BSIFTOP &	  75.07&	 83.46&	  71.23&	 82.46&	  72.37&	 81.67&	  79.84&	 78.01&	  75.83\\ \hline
LPQTOP  &	  69.67&	 78.21&	  82.54&	 71.71&	  83.30&	 78.57&	  83.91&	 78.27&	  76.02\\ \hline
LBPTOP  &	  80.47&	 77.31&	  70.50&	 78.29&	  72.37&	 84.40&	  71.50&	 76.41&	  72.82\\ \hline \hline
DeepFeat&\bf88.92&\bf92.82&\bf88.47&\bf90.24&\bf85.69&\bf89.70&\bf92.69&\bf89.79&\bf88.16\\ \hline
\end{tabular}
\end{center}
\vspace{-.3cm}
\caption{Accuracy (in $\%$) of kinship verification using spatio-temporal and deep features on UvA-NEMO Smile database.}
\label{tab:DeepShallow}
\end{table*}
%---------------------------------
\begin{figure*}[htb!]
\centering
\subfigure[Sister-Sister]{\includegraphics[width=0.246\textwidth]{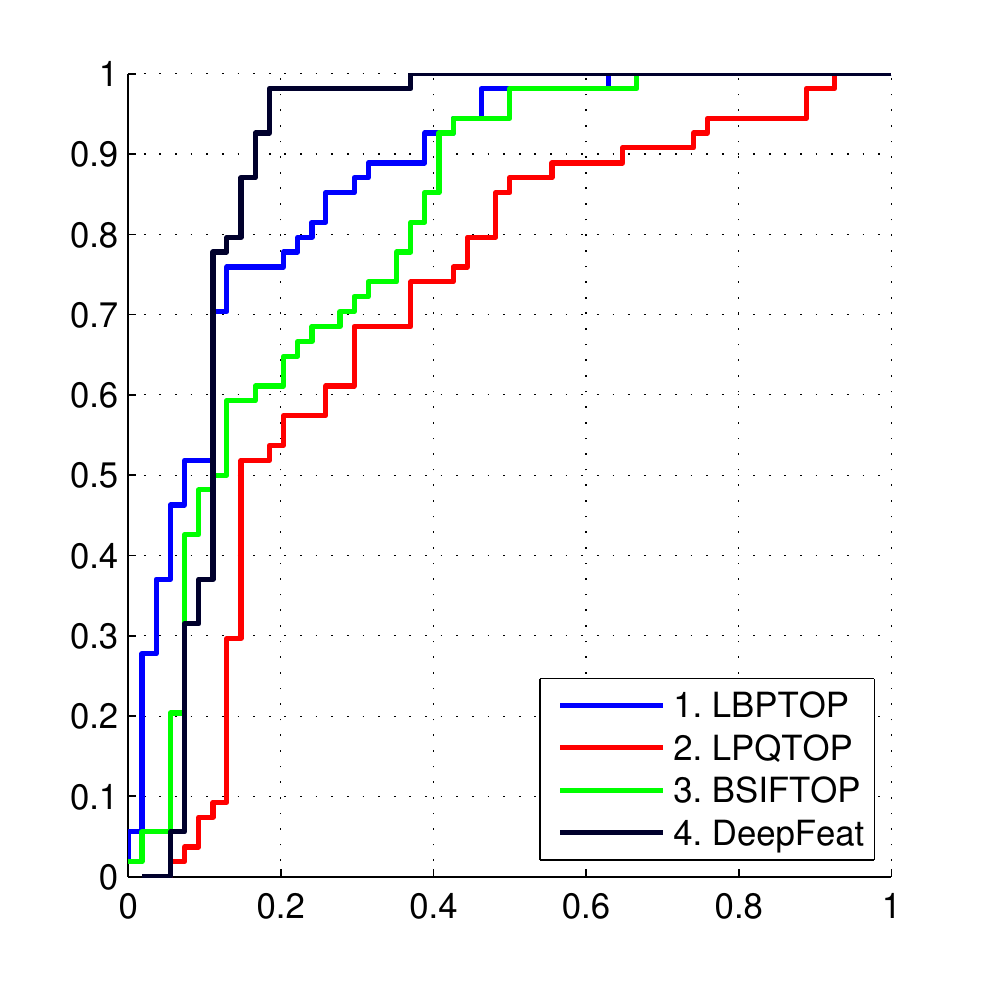}}
\subfigure[Brother-Brother]{\includegraphics[width=0.246\textwidth]{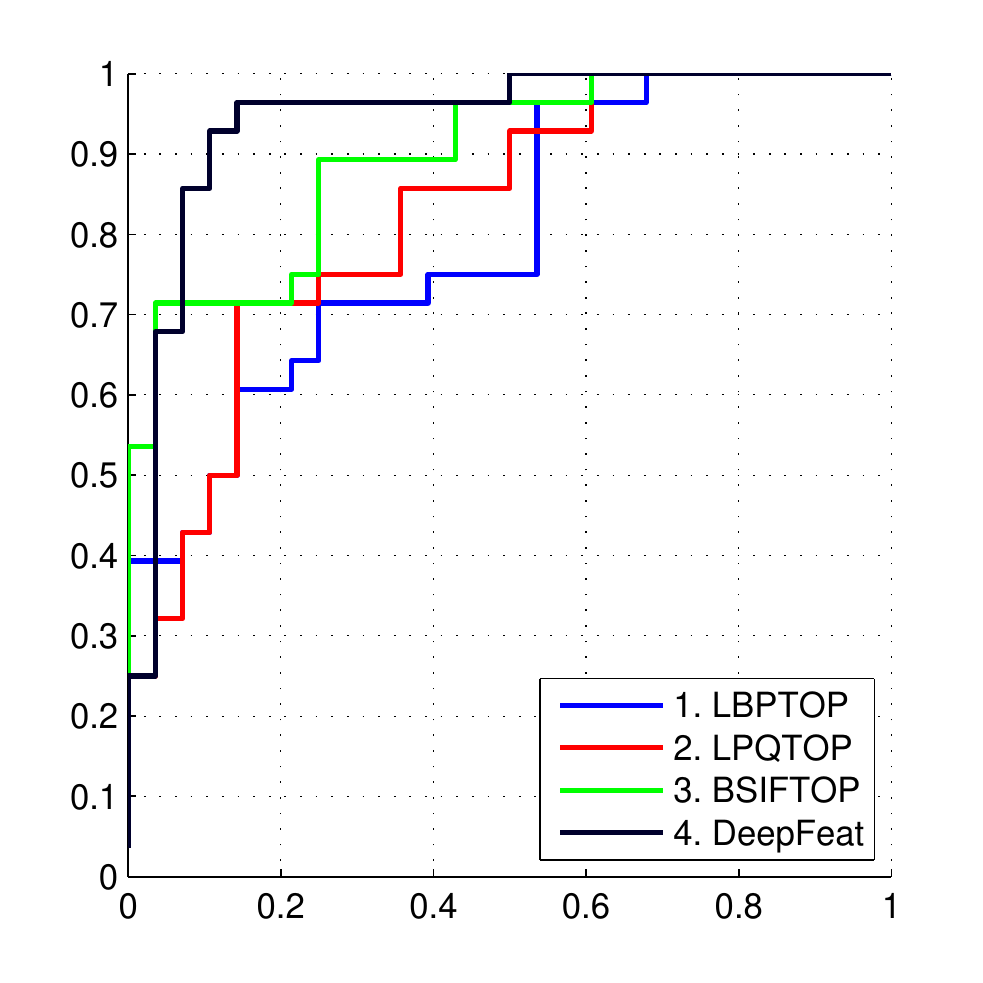}}
\subfigure[Sister-Brother]{\includegraphics[width=0.246\textwidth]{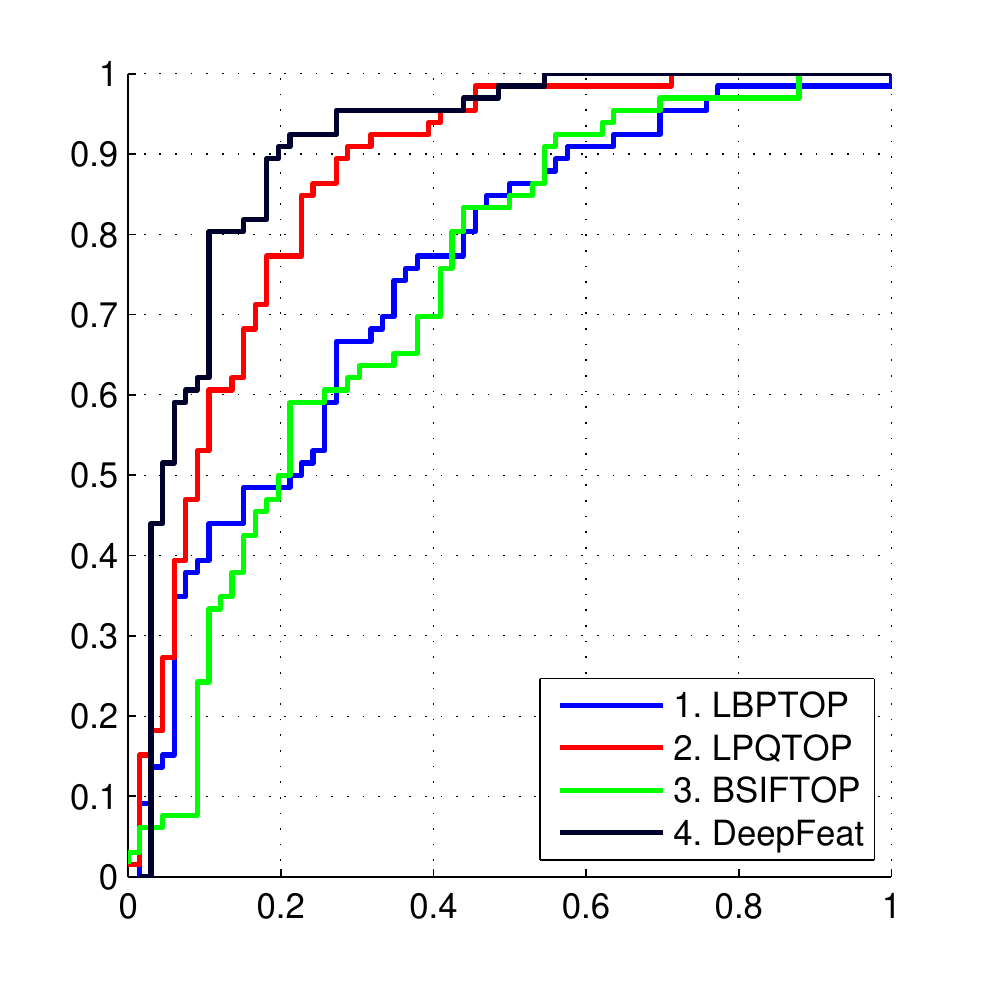}}
\subfigure[Mother-Daughter]{\includegraphics[width=0.246\textwidth]{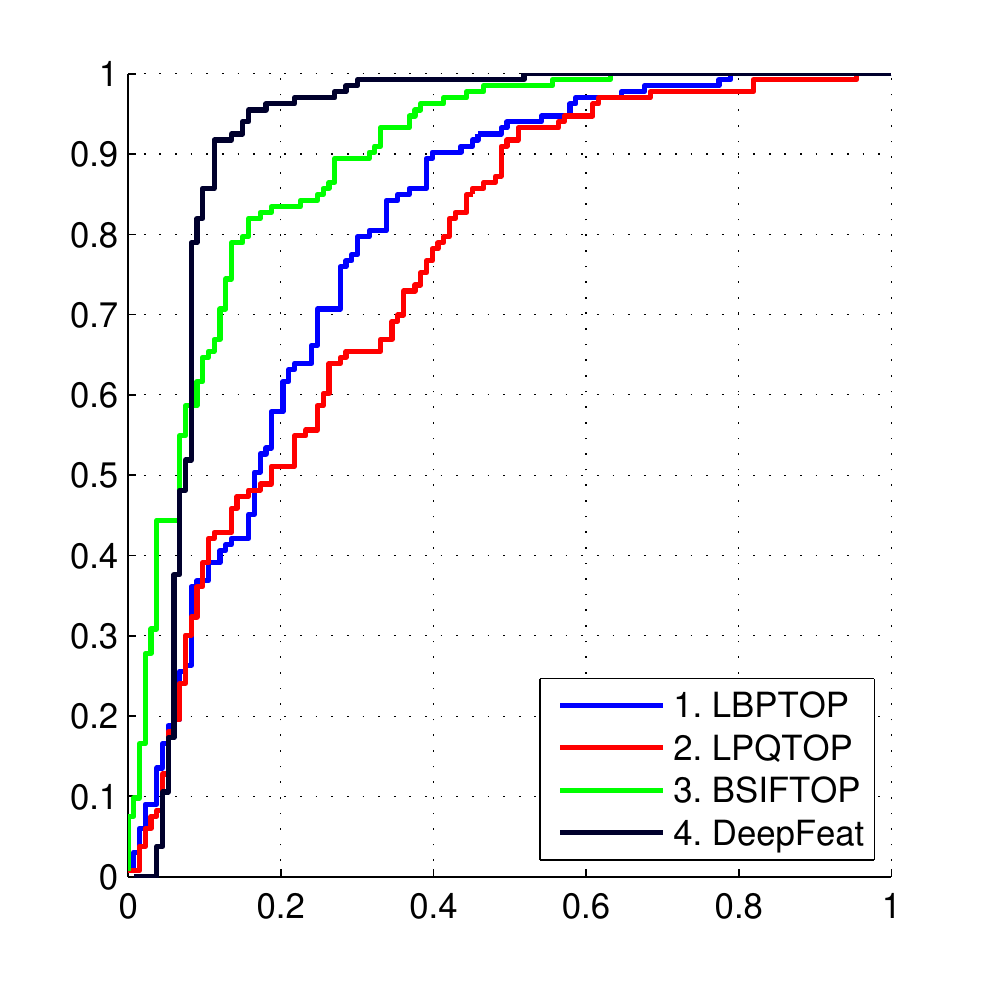}}
\subfigure[Mother-Son]{\includegraphics[width=0.246\textwidth]{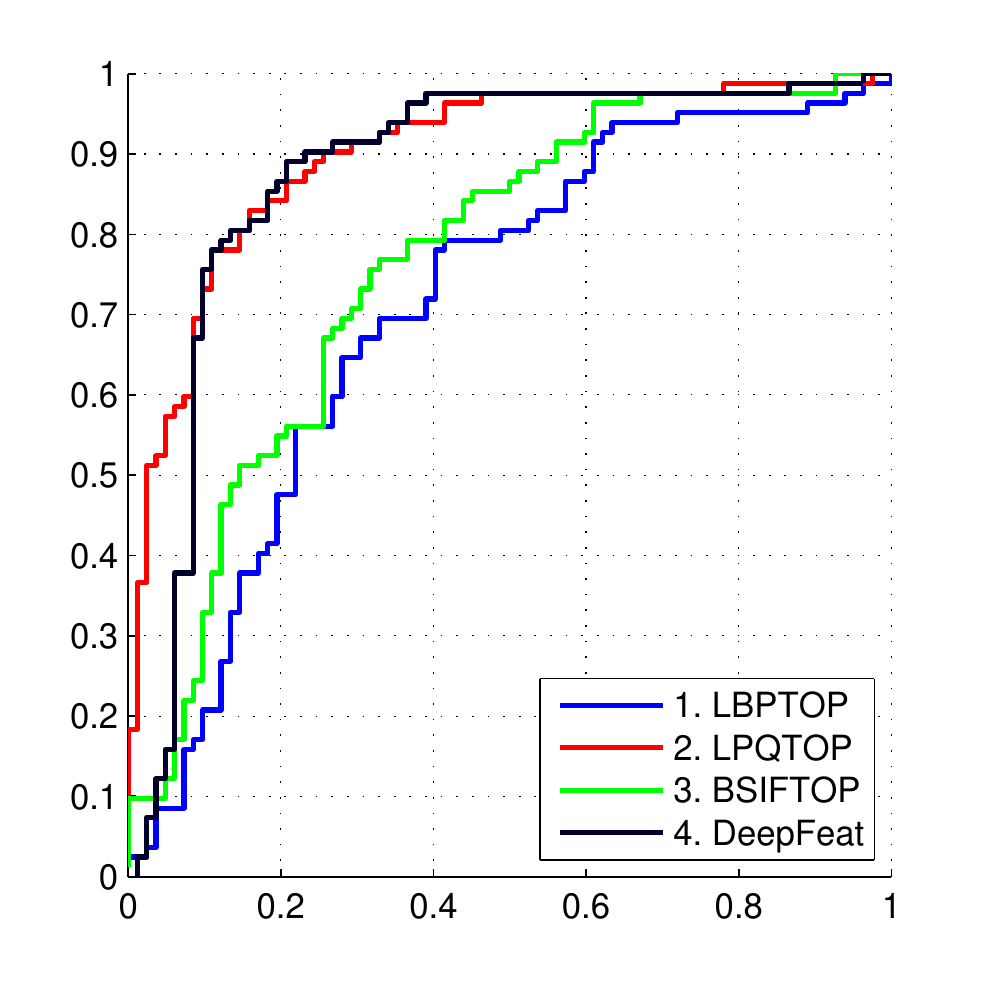}}
\subfigure[Father-Daughter]{\includegraphics[width=0.246\textwidth]{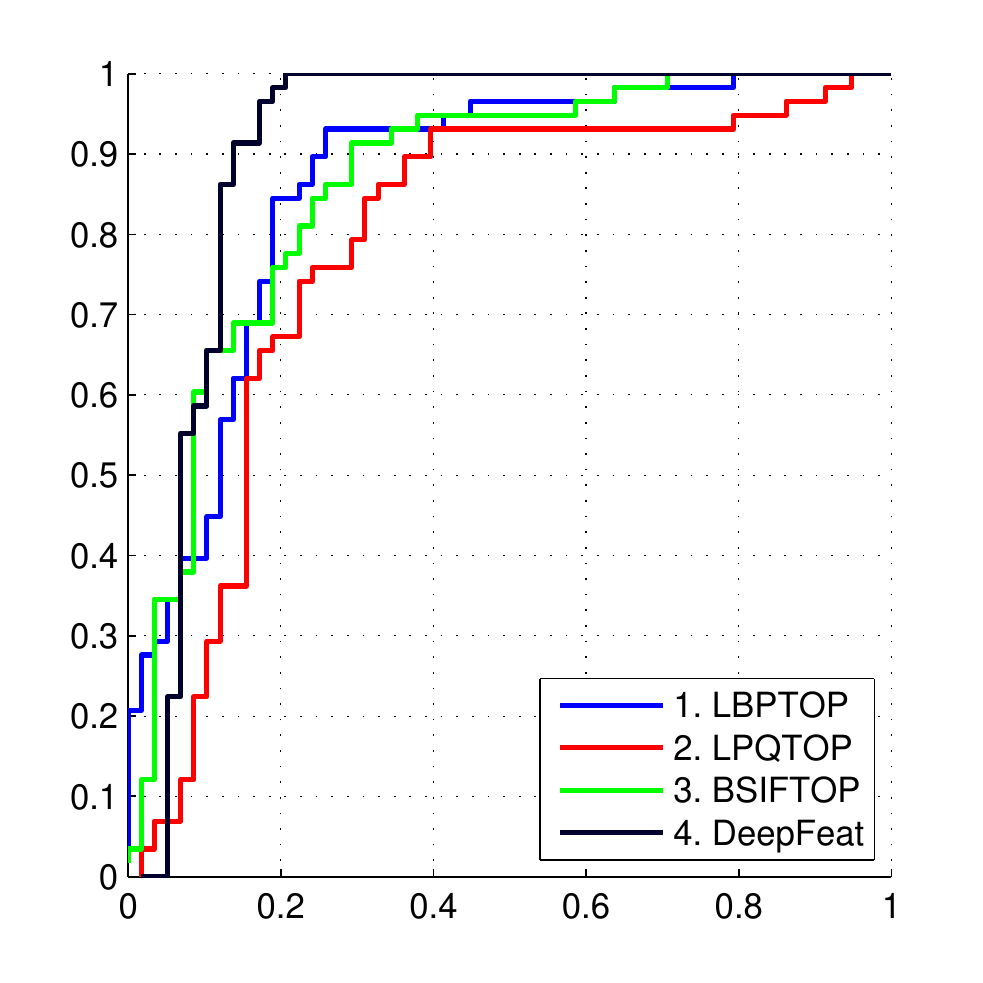}}
\subfigure[Father-Son]{\includegraphics[width=0.246\textwidth]{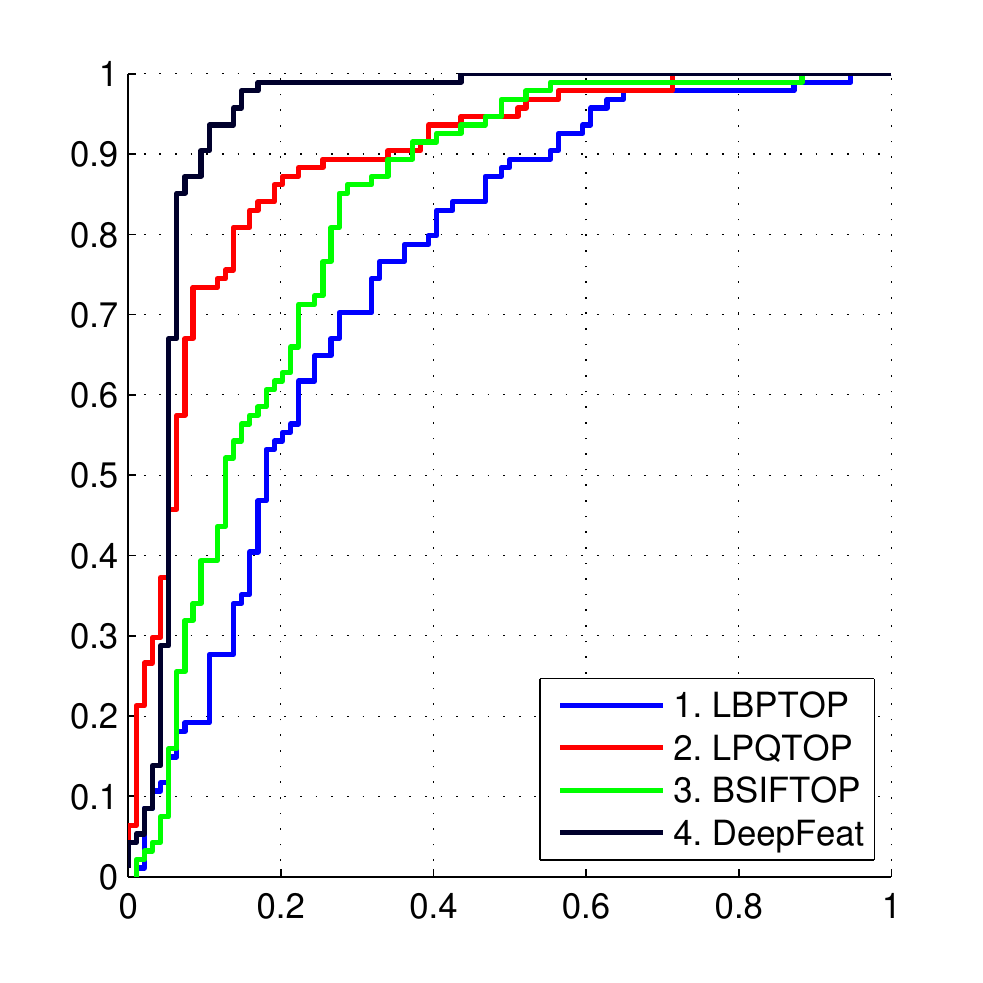}}
\subfigure[All]{\includegraphics[width=0.246\textwidth]{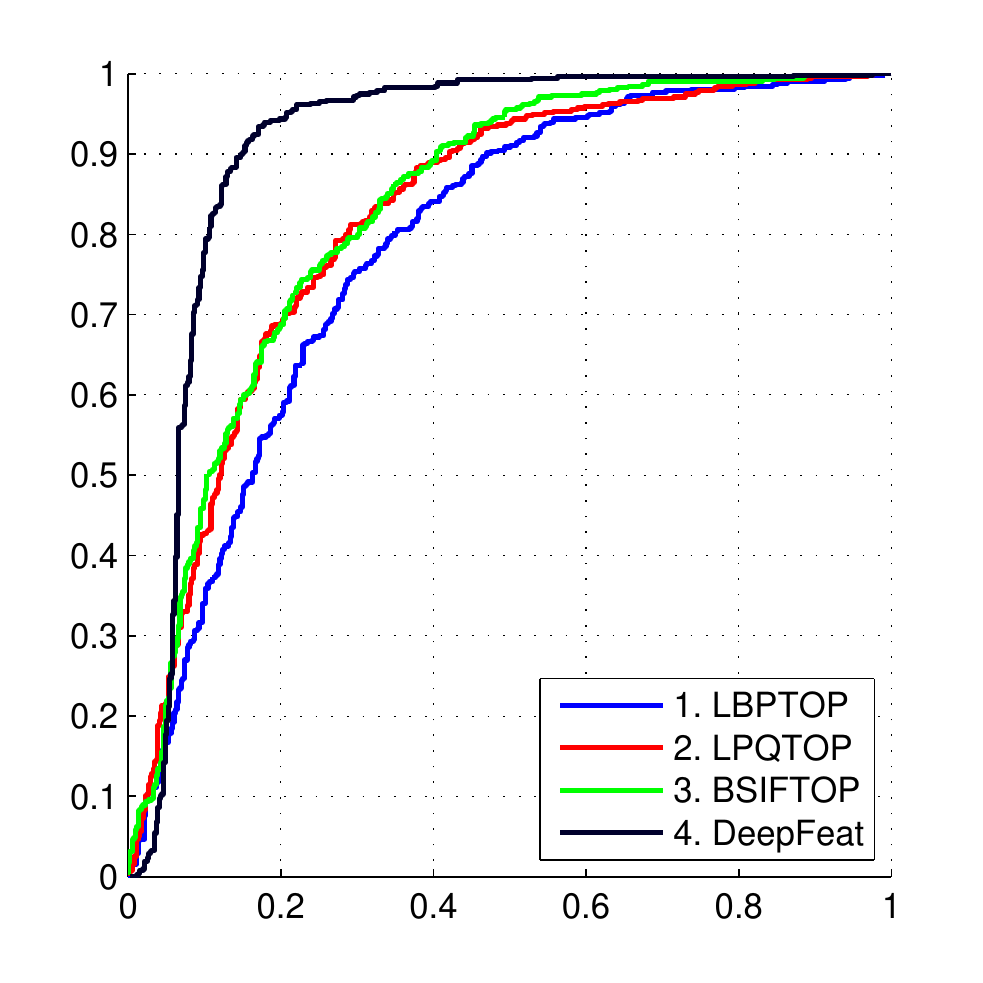}}
\caption{Comparing deep vs. shallow features on UvA-NEMO Smile database.}
\label{fig:DeepShallow}
\end{figure*}
%---------------------------------
Following~\cite{Dibeklioglu_ICCV13}, we randomly generate negative kinship pairs corresponding to each positive pair. Therefore, for each positive pair we associate the first video with a video of another person within the same kin subset while ensuring there is no relation between the two subjects. %Examples of the positive pairs and the generated negative pairs are illustrated by Fig.~\ref{fig:pairsSamples}. 
For all the experiments, we perform a per-relationship evaluation and report the average of spontaneous and posed videos. The accuracy for the whole database, by pooling all the relations, is also provided. Since the number of pairs of each relation is small, we apply a \textit{leave-one-out} evaluation scheme.
%-----------------------------------
\subsection{Results and analysis}
We have performed various experiments to assess the performance of the proposed approach. In the following, we present and analyze the reported results.  

\textit{Deep features against shallow features:}
First we compare the performance of deep features against the spatio-temporal features. The results for different features are reported in Table~\ref{tab:DeepShallow}. The ROC curves for separate relations as well as for the whole database are depicted in Fig.~\ref{fig:DeepShallow}. The performances of the three spatio-temporal features (LBPTOP, LPQTOP and BSIFTOP) show competitive results on different kinship relations. Considering the average accuracy and the accuracy of the whole set, LPQTOP is the best performing method, closely followed by the BSIFTOP, while LBPTOP shows the worst performance.

On the other hand, deep features report the best performance on all kinship relations significantly improving the verification accuracy. The gain in verification performance of the deep features varies between $2\%$ and $9\%$, for different relations, when compared to the best spatio-temporal accuracy. These results highlight the ability of CNNs in learning face descriptors. Even though the network has been trained for face recognition, the extracted face deep features are highly discriminative  when used in the kinship verification task. 

\textit{Comparing relations:}
The best verification accuracy is obtained for B-B and F-S while the lowest are S-B and M-S. These results are maybe due to the different sex of the pairs. One can conclude that checking the kinship relation is easier between persons of the same gender. However, a further analysis of this point is needed as the accuracy of S-S is average in our case. It is also remarkable that the performance of kinship between males (B-B and F-S) is better than between females (M-D and S-S). Moreover, large age differences between the persons composing a pair have an effect on the kinship verification accuracy. For instance, the age difference of brothers (best performance) is lower than it is for M-S (lowest performance).
%---------------------------------
\begin{figure}[tb]
\begin{center}
   \includegraphics[width=.8\linewidth]{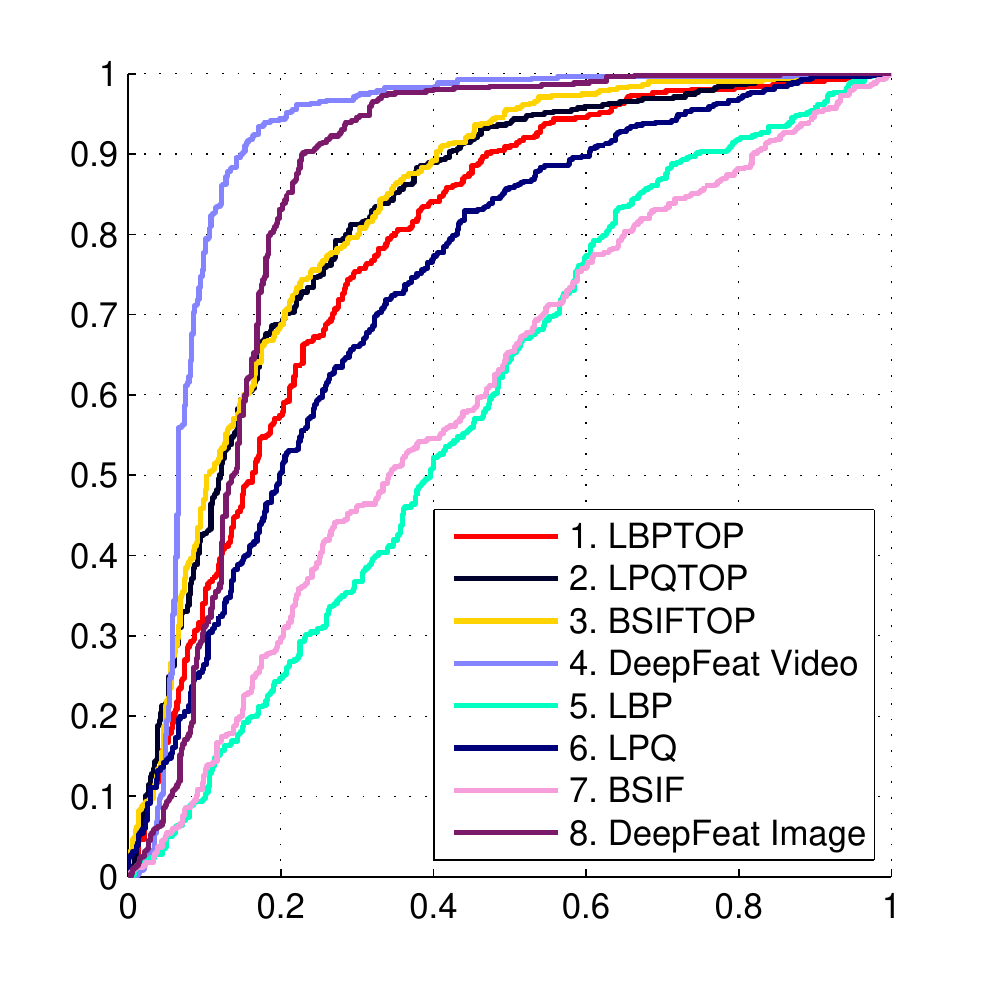}
\end{center}
\vspace{-.3cm}
   \caption{Comparing videos vs. still images for kinship verification on UvA-NEMO Smile database.}
\label{fig:StillVsVideo}
\end{figure}
%-------------------------------------
%%---------------------------------
\begin{table*}[tbh!]
\begin{center}
\begin{tabular}{|l|c|c|c|c|c|c|c||c|c|}
\hline
	Method&S-S& B-B& S-B& M-D& M-S& F-D& F-S & Mean & Whole set \\ \hline
	Fang \textit{et~al.}~\cite{Fang_icip10:1stKin} &61.36 &56.67 &56.25 &56.14 &55.56 &57.14 &55.26& 56.91 &53.51 \\ \hline
	Guo \& Wang~\cite{Guo_TIM12} &65.91& 56.67& 60.94 &58.77 &62.50 &67.86 &55.26 &61.13 &56.14 \\ \hline
	Zhou~\textit{et~al.}~\cite{Zhou_MM12:Gabor_pyramid} &63.64 &70.00& 60.94& 57.02& 56.94& 66.07& 60.53& 62.16& 58.55\\ \hline
	Dibeklioglu~\textit{et~al.}~\cite{Dibeklioglu_ICCV13}&75.00 &70.00 &68.75& 67.54& 75.00 &75.00& 78.95& 72.89 &67.11 \\ \hline \hline
	Our DeepFeat    &   88.92&	 92.82&	  88.47&	 90.24&	  85.69&	89.70&\bf92.69&	  89.79&	 88.16\\ \hline
	Our Deep+Shallow&\bf88.93&\bf94.74&\bf90.07&\bf91.23&\bf90.49&\bf93.10&  88.30&\bf90.98&\bf88.93\\ \hline
\end{tabular}
\end{center}
%\vspace{-.3cm}
\caption{Comparison of our approach for kinship verification against state of the art on UvA-NEMO Smile database.}
\label{tab:soa}
\end{table*}
%--------------------------------------------------------------------------------------------------------------------------------
\textit{Videos vs. images:}
We have carried out an experiment to check if verifying kinship relations from videos instead of images is worthy. Therefore, we employ the first frame from each video of the database. For this experiment, spacial variants of texture features (LBP, LPQ and BSIF) and deep features are extracted from the face images. Fig.~\ref{fig:StillVsVideo} shows the ROC curve comparing the performance of videos against still images for the pool of all relationships. The superiority of the performance of videos compared with still images is obvious for each feature, demonstrating the importance of face dynamics in verifying kinship between persons. Again, deep features extracted from still face images demonstrate high discriminative ability, outperforming both the spatial texture features extracted from images and the spatio-temporal features extracted from videos. We note that, in still images (see Fig.~\ref{fig:StillVsVideo}), LPQ features outperforms both LBP and BSIF, achieving analogous results to the ones computed using video data.

\textit{Feature fusion and comparison against state of the art:} In order to check their complementarity, we have fused spatio-temporal features and deep features. We performed preliminary experiments and empirically found that score-level fusion performs better than feature fusion. In this context, and for simplicity, we have opted for a simple sum at the score-level to perform the fusion. Table~\ref{tab:soa} shows a comparison of the fusion results with the previous works. Overall, the proposed fusion scheme imrpoved further the verification accuracy by a significant margin. This effect is more evident in the relationships depicted by different sex and higher age variation, such as M-S (improved by $4.8\%$) and F-D (improved by $3.4\%$). %As indication, Fig.~\ref{fig:goodClass} and Fig.~\ref{fig:badClass} provide some examples of positive pairs correctly classified and positive pairs wrongly classified by our fusion approach, respectively.

Comparing our results against the previously reported state-of-the-art demonstrates considerable improvements in all the kinship subsets, as shown in Table~\ref{tab:soa}. Depending on the relation type, the improvement in verification accuracy of our approach compared with the best performing method presented by Dibeklioglu~\textit{et~al.}~\cite{Dibeklioglu_ICCV13} ranges from $9\%$ to $23\%$. The average accuracy of all the kin relations has been improved by over $18\%$.  
%%---------------------------------
%\begin{figure}[tbh!]
%\begin{center}
   %\includegraphics[width=\linewidth]{goodClass}
%\end{center}
%\vspace{-.5cm}
   %\caption{Examples of correctly classified positive kin pairs by our approach.}
%\label{fig:goodClass}
%\end{figure}
%%---------------------------------
%%\vspace{-.7cm}
%%---------------------------------
%\begin{figure}[!h]
%\begin{center}
   %\includegraphics[width=\linewidth]{badClass}
%\end{center}
%\vspace{-.5cm}
   %\caption{Examples of wrongly classified positive kin pairs by our approach.}
%\label{fig:badClass}
%\end{figure}

%\vspace{-.5cm}
\section{Conclusion} \label{conc}
In this work, we have investigated the kinship verification problem from face video sequences. In our experiments, faces are described using both spatio-temporal features and deep learned features. Experimental evaluation has been performed using the kinship protocol of the UvA-NEMO Smile video database. We have shown how in kinship verification, the approaches using video information out-perform the ones using still images. Our study demonstrates the high efficiency of face recognition related deep features in describing faces for inferring kinship relations. Further fusion of spatio-temporal features and deep features exhibited interesting improvements in the verification accuracy. A comparison of our approach against the previous state-of-the-art work indicates significant improvements in verification accuracy. 

Even using a pre-trained CNN for face recognition, we have obtained improved results for kinship verification. This demonstrates the generalization ability of deep features to similar tasks. Even though the deep features experiments shown in our work are very promising, these features are extracted in a frame basis way. Employing a video deep architecture would probably lead into better results. However, the scarcity of kinship videos in currently available databases prevented us from opting to such a solution. In this context, future work includes the collection of a large kinship video database including real world challenges to enable learning deep video features. 
%-----------------------------------------------------------------------------------------------------------------------------------
\section*{Acknowledgment} 
E. Boutellaa is acknowledging the financial support of the Algerian MESRS and CDTA under the grant number 060/PNE/ENS/FINLANDE/2014-2015. The support of the Academy of Finland, Northwestern Polytechnical University and the Shaanxi Province is also acknowledged.
%-----------------------------------------------------------------------------------------------------------------------------------
{\small
\bibliographystyle{ieee}
\bibliography{egbib}
}
\newpage%-----------------------------------------------------------------------------------------------------------------------------------
\end{document}